\documentclass{./styles/svproc}

\usepackage{url}
\usepackage{xcolor}
\usepackage{todonotes}
\usepackage{tcolorbox}
\usepackage{svg}
\usepackage{tabularx}
\usepackage{cite}  

\begin{document}
\mainmatter              
%
%
\title{Addressing Failures in Robotics using Vision-Based Language Models (VLMs) and Behavior Trees (BT)\thanks{This project is supported by WASP (Wallenberg AI, Autonomous Systems and Software Program).}}
\titlerunning{Addressing Failures in Robotics}  
%
\author{Faseeh Ahmad\inst{1}, Jonathan Styrud\inst{2} \and
Volker Krueger\inst{1}}
\authorrunning{Faseeh Ahmad et al.} 
%
\tocauthor{Faseeh Ahmad, Jonathan Styrud and Volker Krueger}
\institute{Lund University, Lund, Sweden,\\
\email{\{faseeh.ahmad, volker.krueger\}@cs.lth.se}
\and
KTH Royal Institute of Technology, Stockholm, Sweden\\
\email{jstyrud@kth.se}}

\maketitle              
\vspace{-0.5cm}
\begin{abstract}
In this paper, we propose an approach that combines Vision Language Models (VLMs) and Behavior Trees (BTs) to address failures in robotics. Current robotic systems can handle known failures with pre-existing recovery strategies, but they are often ill-equipped to manage unknown failures or anomalies. We introduce VLMs as a monitoring tool to detect and identify failures during task execution. Additionally, VLMs generate missing conditions or skill templates that are then incorporated into the BT, ensuring the system can autonomously address similar failures in future tasks. We validate our approach through simulations in several failure scenarios. 
\end{abstract}
\vspace{-1.0cm}
\section{Introduction}

Modern robotic systems can handle complex tasks in controlled environments, but transitioning into dynamic, small-batch manufacturing introduces new challenges, especially around failure management. Failures; unforeseen disturbances that prevent task completion; can lead to costly delays and risks, particularly in shared workspaces~\cite{wu2021error}. The ability to detect, identify, and recover from failures autonomously is crucial for ensuring the robustness of robotic systems.

Traditional failure management strategies in robotics include human intervention, failure analysis~\cite{jusuf2021defense}, and automated recovery strategies~\cite{wu2021error}. These approaches have limitations: human intervention is time-consuming, failure analysis requires expertise, and automated strategies often lack flexibility in handling unforeseen scenarios. Our recent work~\cite{ahmad2024recovery} introduced a novel method using automated recovery behaviors modeled as robotic skills with parameters, preconditions, and postconditions, executed through Behavior Trees and Motion Generators (BTMG) policy representation \cite{rovida2017bt}.  This approach optimizes recovery policies using Reinforcement Learning (RL)~\cite{skills2022} and also adapts the parameters to different task variations~\cite{ahmad2023}.

However, two key limitations remain: (1) the system assumes failure detection and identification are already solved, requiring prior knowledge of the failure, and (2) it only handles known failures with predefined solutions. These limitations make it difficult to address unforeseen failures. We propose addressing these gaps by utilizing Vision Language Models (VLMs) to detect, identify, and generate solutions for unknown failures. By integrating VLMs with Behavior Trees (BTs), our approach autonomously monitors task execution, identifies failure states, and generates missing conditions or skill templates to recover from failures. The BT is then updated using a reactive planner~\cite{styrud2024betrxp} to handle similar future occurrences.

\vspace{-0.3cm}
\subsection*{Main Contributions}
\begin{itemize}
    \item We propose a novel integration of VLMs with BTs for monitoring, failure detection, identification and recovery in robotic systems.
    \item We use VLMs to generate missing preconditions or skill templates to address failures and update the BT policy.
    \item We conduct experiments to demonstrate the effectiveness of the approach.
\end{itemize}

\section{Background}
This section provides essential background concepts to our proposed approach, focusing on behavior trees, reactive planner and vision-based language models.
\vspace{-0.3cm}
\subsection{Behavior Trees (BT)}
Behavior Trees (BTs) are hierarchical models for task execution, known for their modularity and flexibility \cite{colledanchise2018bt}. A BT organizes task execution through nodes that receive tick signals, indicating readiness for execution. BTs consist of two types of nodes: control-flow nodes and execution nodes. \textit{Control-flow nodes} manage execution flow and return statuses of success, failure, or running; examples include Sequence (AND) and Fallback/Selector (OR). \textit{Execution nodes}, which are leaf nodes, represent either skills ("!") or conditions ("?"). Skills perform specific tasks, while conditions evaluate the environment, returning only success or failure. BTs offer modularity and clarity, making them ideal for robotics applications, particularly in dynamic environments where flexibility is required \cite{iovino2022btmodularity}.
\vspace{-0.3cm}

\subsection{Reactive Planner}
Reactive planners generate BTs dynamically, using a backchaining approach to select skills that satisfy goal conditions~\cite{colledanchise2019blended}. The process iteratively selects skills based on their preconditions and postconditions, constructing a BT that satisfies the specified goal. This recursive process continues until a leaf node is reached or a predefined depth is attained. The planner ensures adaptability by dynamically responding to changes in the environment without requiring extensive re-planning. This planner has been extended for various applications in~\cite{iovino2023framework,styrud2024bebop}.
\vspace{-0.3cm}
\subsection{Vision Language Models (VLM) in Robotics} Vision Language Models (VLMs) are powerful tools in robotics, enabling a deeper understanding of complex environments by combining visual data with language inputs. VLMs excel at tasks such as scene interpretation, object recognition, and generating control skills based on visual cues and task descriptions \cite{chen2024recovery, OpenVLA2024}. Recent applications of VLMs in robotics include failure recovery, task planning, and multimodal reasoning, with systems like ReplanVLM~\cite{Mei2024ReplanVLM} and AHA~\cite{AHA2024} demonstrating their ability to reason over failures and generate dynamic solutions.

\section{Approach}
We extend the existing framework~\cite{ahmad2024recovery} to handle unknown failures by integrating VLMs for failure detection, identification and recovery and generate missing preconditions or skill templates to be incorporated into the BT.
\vspace{-0.5cm}
\begin{figure}
    \centering
    \includegraphics[width=1.0\linewidth]{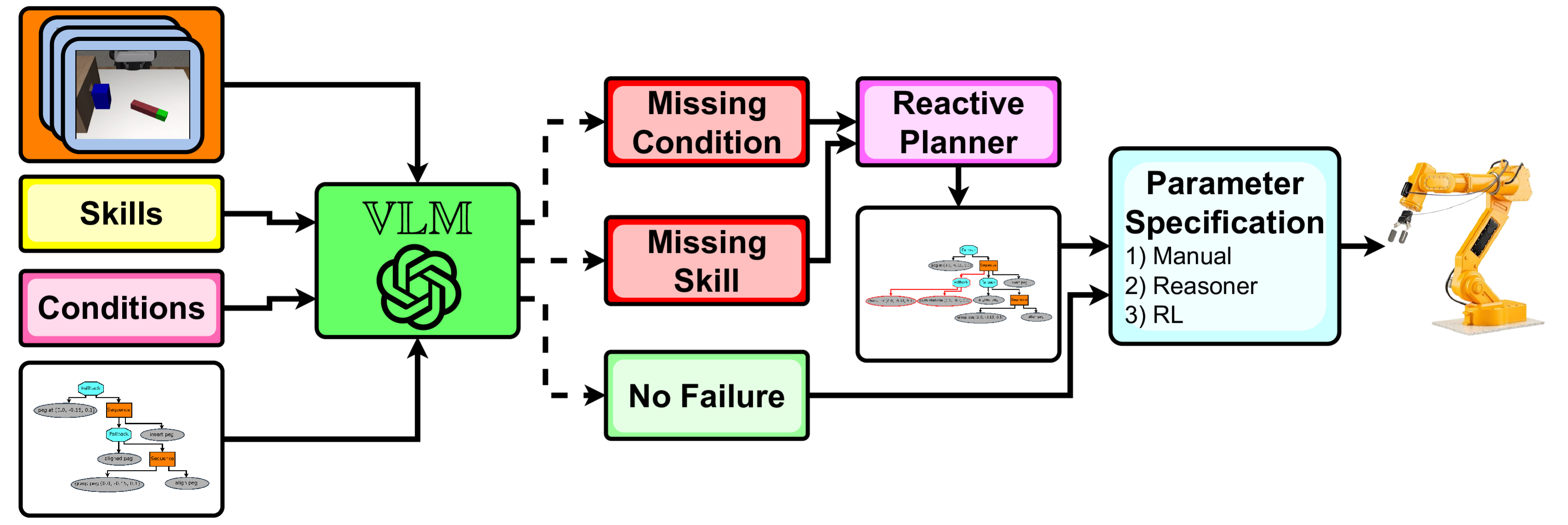}
    \caption{Overview of the proposed approach, where the VLM takes a set of images, skills, conditions, and a BT as input. The VLM uses this information to provide missing conditions or skills, which are then used to update the BT through a planner.}
    \label{fig:appraoch}
\end{figure}
\vspace{-1.0cm}

\subsection{Failure Detection and Idenitification}
\begin{figure}[h]
    \centering
    \includegraphics[width=0.8\linewidth]{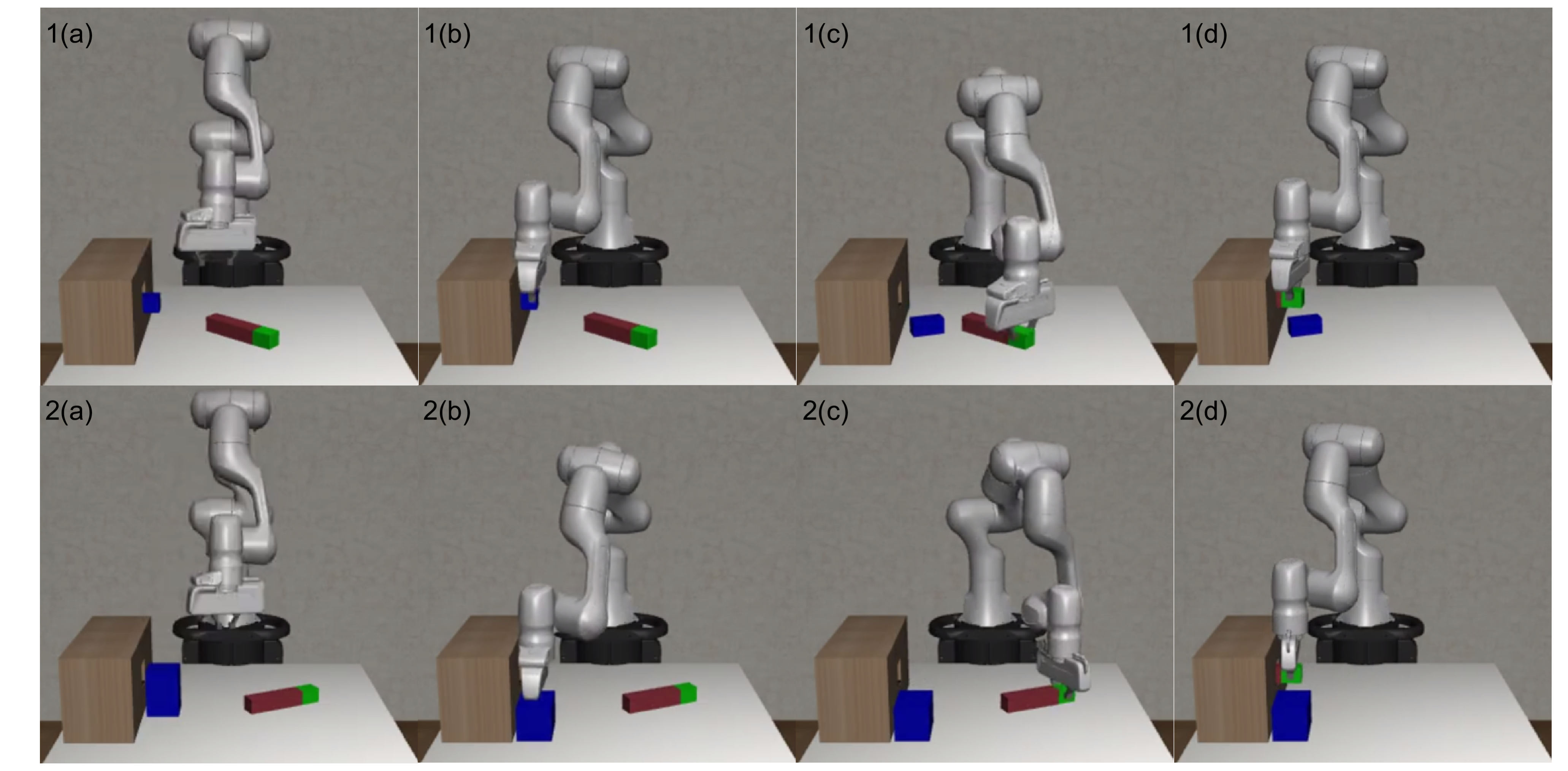}
    \caption{Scenes showing peg-in-hole task execution with obstacles. The first row (1(a)–1(d)) illustrates the task with a small obstacle, while the second row (2(a)–2(d)) depicts the task with a large obstacle.}
    \label{fig:scenes}
    \vspace{-0.5cm}
\end{figure}

Failure detection refers to the system's ability to recognize when a task cannot be completed due to unforeseen errors, such as hardware malfunctions or environmental disturbances. This can be achieved by using sensor data, such as from cameras or force-torque sensors, and comparing it against the expected postconditions of skills. For example, in a peg-in-hole task, if an object blocks the hole, the system detects this failure when the postcondition of the "insert" skill (peg inserted) is not met \cite{wu2021error} (see Figure~\ref{fig:scenes}).

Failure identification involves describing the failure using existing system conditions and understanding why the task could not be completed. For instance, in the peg-in-hole task, the missing precondition when an obstacle is blocking the hole could be "Not any obstacle at hole" for the insert skill. This allows the system to formulate strategies for dealing with similar failures in the future.    
\vspace{-0.5cm}
\begin{figure}[h]
    \centering
    \includegraphics[width=1.0\linewidth]{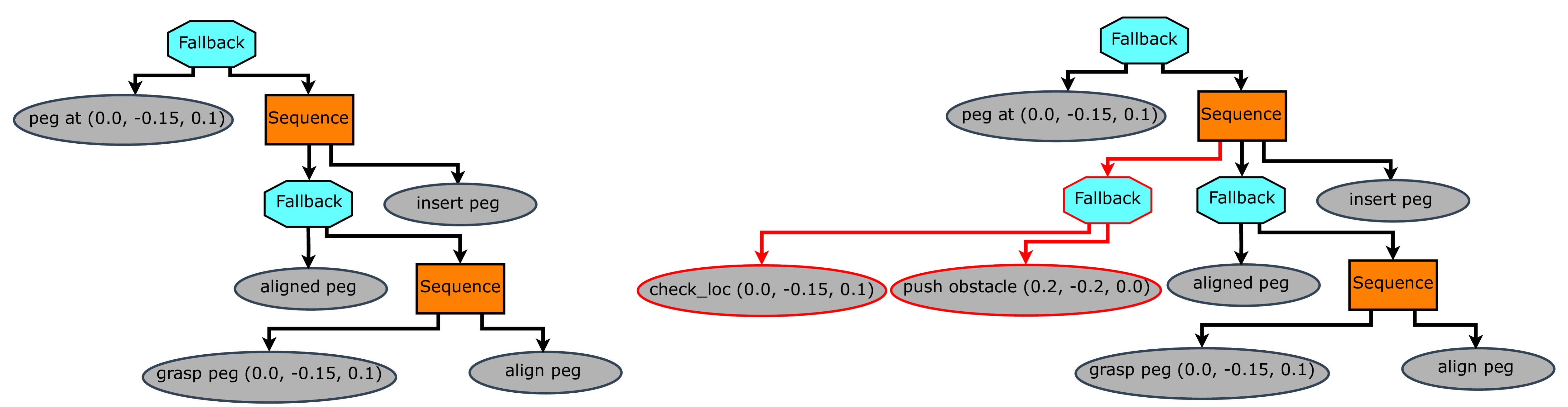}
    \caption{Comparison of Behavior Trees (BTs). The left side shows the initial BT, while the right side illustrates the updated BT, with the changes highlighted in red connections.}
    \label{fig:bt_one}
\end{figure}
\subsection{Monitoring using VLM}
\vspace{-0.1cm}
We use VLMs to enable failure detection, identification and recovery (Figure~\ref{fig:appraoch}). Before task execution, the VLM is queried with images of the task environment, the BT structure, and the skills and conditions involved. The VLM then assesses whether the task will succeed and, if not, identifies the cause of failure (detection). It also suggests the missing condition (identification) that could prevent the failure. If the system lacks the required skill to recover from a failure (recovery), the VLM suggests an appropriate recovery skill based on the provided library of existing skills (see Figure~\ref{fig:bt_one}).

\subsection{Condition and Skill Template Generation}
When the VLM identifies a missing condition or skill, it updates the BT accordingly. For example, if a small obstacle blocks the peg, the condition "hole free" is generated and added as a precondition to the "insert" skill. The reactive planner regenerates the BT by incorporating this condition, ensuring similar failures are handled in the future \cite{styrud2024betrxp}. If a recovery skill is missing, the VLM generates an skill template that follows a structured format and requires some manual inputs to complete. For instance, if a large object blocks the peg hole and the gripper cannot grasp it, the VLM suggests a "push" skill template to remove the obstacle (see Figure~\ref{fig:scenes}). This template is added to the BT, and in the future, this process could be fully automated, allowing the system to autonomously recover from failures.

\section{Experiments}
We validated the proposed approach using simulations in robosuite~\cite{robosuite2023} and OpenAI's GPT-4. The experiments were designed around three tasks, each involving unknown failures:
\begin{itemize}
    \item \textit{Peg-in-Hole Task:} Two scenarios—(A) a small obstacle inside the hole, and (B) a large obstacle positioned in front of the hole.
    \item \textit{Lift Task:} An additional cube is placed on top of the target object, creating an unforeseen failure.
    \item \textit{Door Opening Task:} The robot attempts to open a door but lacks the precondition that the handle must be turned first.
\end{itemize}

\section{Evaluation Metrics and Results}
We evaluated the VLM's performance using three key metrics: consistency in failure detection and recovery, the importance of vision input, and skill feasibility considerations (ensuring that suggested skills, such as a "grasp" skill, are feasible based on the gripper's affordance and object location). For all experiments, we used model parameters of temperature and top\_p set to 0.1, which resulted in more deterministic and focused outputs, reducing randomness and ensuring that the model consistently chose the most likely responses.

\begin{itemize}
    \item \textbf{Consistency of Failure Detection and Recovery:} The VLM's reliability was tested across 10 trials per scenario, consistently detecting and recovering from failures in all tasks, achieving 100\% consistency.

    \item \textbf{Vision Importance Ablation Study:} To assess the impact of visual input, we compared VLM (with visual input) and LLM (without visual input). In the \textit{Peg-in-Hole (small obstacle)}, \textit{Lift}, and \textit{Door Opening} tasks, both models achieved 100\% accuracy. However, in the \textit{Peg-in-Hole (large obstacle)} task, the VLM achieved 100\% accuracy, while the LLM reached 30\% accuracy without skill feasibility considerations and 60\% with feasibility checks.

    \item \textbf{Skill Feasibility Considerations:} When skill feasibility is considered, LLM performance improved but still fell short of the VLM. The VLM excelled in complex scenarios like the \textit{Peg-in-Hole (large obstacle)} task, generating feasible recovery skills autonomously.
\end{itemize}

\section{Conclusion and Future Work}
In this paper, we introduced a method for integrating Vision Language Models (VLMs) with Behavior Trees (BTs) to autonomously detect, identify, and recover from failures in robotic systems. By generating missing conditions and skill templates, the system can effectively handle unknown failures and adapt its execution policy for future tasks.
Future work will focus on several key areas: expanding the range of failure scenarios to include more complex and dynamic environments, improving the skill generation mechanism to move from generating skill templates to directly producing feasible skills, thereby reducing manual input. Additionally, we aim to fine-tune an open-source model to further enhance the system's performance and adaptability across diverse robotic tasks.


\end{document}